\documentclass[runningheads]{llncs}

\usepackage{eccv}

\usepackage{eccvabbrv}

\usepackage{graphicx}
\usepackage{booktabs}
\usepackage{algorithm}
\usepackage{algorithmic}
\usepackage{multirow}
\usepackage{tcolorbox}
\tcbuselibrary{breakable}

\usepackage{color}
\usepackage{colortbl}
\definecolor{tableBlue}{rgb}{0.424,0.557,0.749}
\definecolor{lightblue}{rgb}{0.85, 0.95, 1}
\definecolor{persiangreen}{rgb}{0.0, 0.65, 0.58}
\definecolor{cadmiumred}{rgb}{0.89, 0.0, 0.13}
\newcommand{\tableLineColor}{\rowcolor{lightblue}}
\definecolor{increaseGreen}{rgb}{0.51,0.702,0.4}
\definecolor{decreaseRed}{rgb}{0.647,0.337,0.318}
\newcommand{\increasingData}[1]{\textcolor{persiangreen}{$\uparrow$#1}}

\usepackage[accsupp]{axessibility}  

\usepackage{hyperref}

\usepackage{orcidlink}

\begin{document}

\title{Thinking in Streaming Video} 

\titlerunning{Thinking in Streaming Video}

\author{
Zikang Liu\inst{1,2}\thanks{Equal contribution.}
\and
Longteng Guo\inst{1}$^{\star}$
\and
Handong Li\inst{1,2}$^{\star}$
\and
Ru Zhen\inst{3}
\and
Xingjian He\inst{1}
\and
Ruyi Ji\inst{1}
\and
Xiaoming Ren\inst{3}
\and
Yanhao Zhang\inst{3}
\and
Haonan Lu\inst{3}
\and
Jing Liu\inst{1,2}\thanks{Corresponding author.}
}

\authorrunning{Z.~Liu et al.}

\institute{
Institute of Automation, Chinese Academy of Sciences
\and
School of Artificial Intelligence, University of Chinese Academy of Sciences
\and
OPPO AI Center, OPPO Inc.
\\
\email{\{liuzikang2023\}@ia.ac.cn, \{longteng.guo,jliu\}@nlpr.ia.ac.cn}}

\maketitle

\begin{abstract}
Real-time understanding of continuous video streams is essential for interactive assistants and multimodal agents operating in dynamic environments. 
However, most existing video reasoning approaches follow a batch paradigm that defers reasoning until the full video context is observed, resulting in high latency and growing computational cost that are incompatible with streaming scenarios. 
In this paper, we introduce \textbf{ThinkStream}, a framework for streaming video reasoning based on a \textbf{Watch--Think--Speak} paradigm that enables models to incrementally update their understanding as new video observations arrive. 
At each step, the model performs a short reasoning update and decides whether sufficient evidence has accumulated to produce a response. 
To support long-horizon streaming, we propose Reasoning-Compressed Streaming Memory (RCSM), which treats intermediate reasoning traces as compact semantic memory that replaces outdated visual tokens while preserving essential context.
We further train the model using a Streaming Reinforcement Learning with Verifiable Rewards scheme that aligns incremental reasoning and response timing with the requirements of streaming interaction. 
Experiments on multiple streaming video benchmarks show that ThinkStream significantly outperforms existing online video models while maintaining low latency and memory usage. Code, models and data will be released at \url{https://github.com/johncaged/ThinkStream}

\end{abstract}

\section{Introduction}
\label{sec:intro}

\begin{figure*}[t]
\centering
\includegraphics[width=\linewidth]{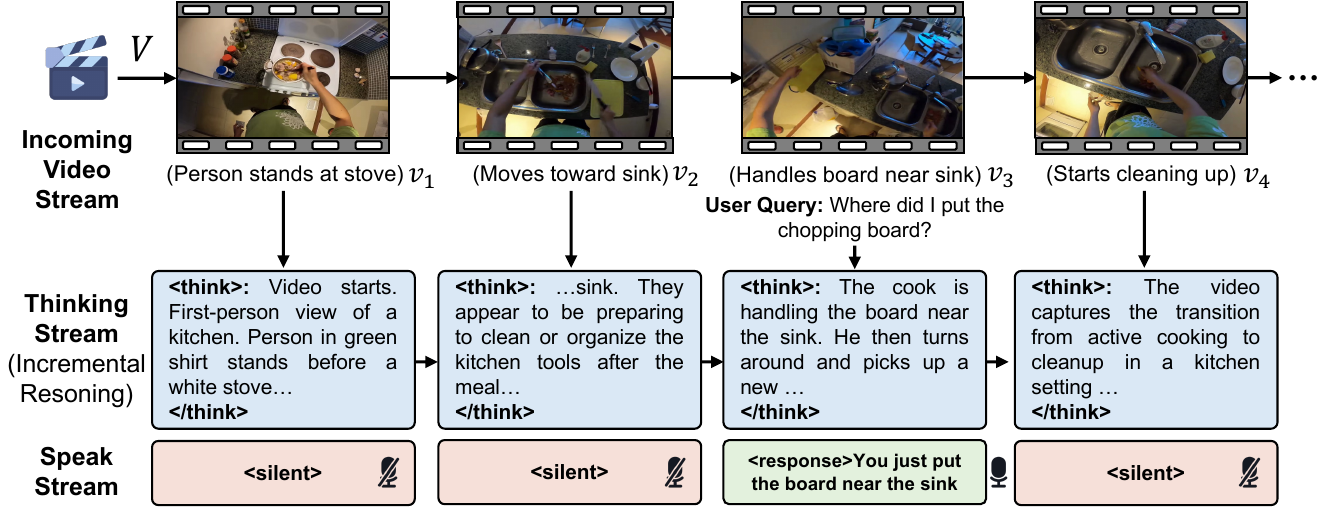}
\caption{
Illustration of the \textbf{Streaming Watch--Think--Speak} paradigm. 
As video chunks arrive sequentially, the model continuously updates its understanding through incremental reasoning steps (\texttt{<think>}). 
Each update integrates newly observed evidence with accumulated context. 
Based on this evolving interpretation, the model decides whether sufficient evidence has been gathered to produce a response (\texttt{<response>}) or whether it should remain silent (\texttt{<silent>}) and continue observing the stream.
}
\label{fig:intro}
\end{figure*}

Real-time understanding of continuous video streams is a core capability for interactive assistants and multimodal agents~\cite{wang2025streambridge,xu2025streamingvlm,xia2025streaming}.
Unlike offline video question answering~\cite{zhang2024llava,li2025breaking,Qwen2.5-VL,bai2025qwen3}, streaming scenarios impose several fundamental constraints: 
(i) \emph{strict causality}, where reasoning must rely only on observations available up to the current moment; 
(ii) \emph{low computation and memory usage}, requiring inference latency and resource usage to remain stable even as the stream grows indefinitely; and 
(iii) \emph{timely interaction}, which demands that the system maintain an up-to-date interpretation of the evolving scene to respond or act when necessary. 

Such constraints commonly arise in environments where visual events unfold continuously, including real-time assistants observing user activities, monitoring systems detecting emerging events, and embodied agents interacting with dynamic physical environments.

However, most existing video reasoning paradigms remain fundamentally \emph{batch}: the model first consumes a long video context and only then performs multi-step reasoning to produce an answer~\cite{wu2025tempr1,feng2025video}. 
This design introduces unacceptable latency and weakens the connection between reasoning steps and the evidence that triggered them. 
Increasing the depth or length of Chain-of-Thought (CoT) reasoning partially mitigates this issue, but long-form decoding further inflates latency and compute, making it incompatible with continuous streams.

In dynamic environments, human cognition often resembles \emph{thinking while observing}: people form provisional interpretations from local evidence, refine them as new signals arrive, and act once sufficient confidence has been accumulated. 
This observation motivates an incremental perspective on streaming video understanding, where reasoning evolves alongside the incoming stream rather than being deferred until the full context is observed.

In this paper, we propose a \textbf{Streaming Watch--Think--Speak} paradigm for video understanding. 
At each step, the model observes a newly arrived video chunk and performs a short reasoning update that integrates the new evidence with previously accumulated context. 
Based on this evolving understanding, the model determines whether the available evidence is sufficient to produce a response, or whether it should remain silent and continue observing the stream until more information arrives. 
As illustrated in ~\cref{fig:intro}, the model produces a structured output consisting of (i) a reasoning segment enclosed in a \texttt{<think>} block and (ii) an interaction action that either emits a response (\texttt{<response>}) or indicates that the model remains silent (\texttt{<silent>}) and continues observing the stream.

Building on this paradigm, we introduce \textbf{ThinkStream}, a framework that enables reasoning-capable multimodal models to operate over long-horizon video streams under low computational resources. 
The central challenge is to preserve coherent understanding without allowing the visual context, and thus the KV cache, to grow rapidly. 
To address this, we propose \textbf{Reasoning-Compressed Streaming Memory (RCSM)}, which treats intermediate reasoning traces as compact semantic memory. 
As the stream evolves, outdated visual tokens are evicted while the corresponding reasoning states are retained as long-term semantic anchors, preserving essential historical context while keeping inference cost stable.

To train models that produce such reasoning updates, we introduce a \textbf{Streaming Reinforcement Learning with Verifiable Rewards (RLVR)} framework tailored to Watch--Think--Speak. 
We design simple rule-based rewards that are automatically verifiable and reflect the requirements of streaming interaction, encouraging the model to generate structured reasoning traces, respond only when sufficient evidence has accumulated, and maintain answer correctness throughout the stream.

Finally, we support large-scale RLVR training and inference with an efficient streaming rollout backend that enables high-throughput inference with dynamic context updates. 
We also construct a large-scale dataset featuring time-grounded reasoning traces for cold-start supervision and a carefully formatted subset for RLVR training.

Extensive experiments validate the effectiveness of our approach. 
On dedicated streaming video benchmarks, ThinkStream achieves substantially stronger performance than existing online video models, with even a compact 3B-scale model outperforming significantly larger baselines. 
Meanwhile, our framework maintains low latency and memory usage over long video streams, enabling real-time inference while preserving strong performance on standard offline video benchmarks.

Our contributions are summarized as follows:
\begin{itemize}
\item We propose a \textbf{Streaming Watch--Think--Speak} paradigm that formulates streaming video understanding as an incremental reasoning and interaction process, enabling models to continuously update their interpretation while deciding when to respond.

\item We introduce \textbf{ThinkStream}, a unified framework for long-horizon streaming video reasoning, together with Reasoning-Compressed Streaming Memory (RCSM) that treats reasoning traces as compact semantic memory to replace evicted visual tokens while keeping inference cost much lower.

\item We develop a \textbf{Streaming Reinforcement Learning} with RLVR training scheme that aligns incremental reasoning and response timing through automatically verifiable reward signals.

\item We support scalable training and deployment through an efficient streaming inference backend and construct a large-scale dataset with time-grounded reasoning traces. 
Our code, models, and datasets will be released to facilitate future research on streaming video reasoning.

\end{itemize}

\section{Related Work}

\subsection{Streaming Video Understanding}

Transitioning to streaming video for real-time interaction, early frameworks (e.g., VideoLLM-online~\cite{chen2024videollm}) rely on direct perception-response cycles lacking intermediate reasoning. To manage massive visual token influxes, subsequent methods introduced specialized memory architectures~\cite{zhang2025flash,zhao2025cogstream,li2025lion}, KV cache compression~\cite{yang2025streammem,zhang2026hermes,qian2024streaming} or KV cache sparsification/pruning~\cite{xu2025streamingvlm,chen2025streamkv}. However, their memory mechanisms remain purely visual, devoid of semantic compression and explicit logical deduction. Furthermore, addressing the proactive output timing ("when to speak") challenge, existing approaches (e.g., EgoSpeak~\cite{kim2025egospeak}, StreamMind~\cite{ding2025streammind}) heavily depend on dedicated classification heads or external event triggers. Conversely, our framework eschews external classifiers; by integrating a generative reasoning phase, the model autonomously determines whether to remain silent or respond, seamlessly unifying perception, reasoning, and reaction into a single generative process.

\subsection{Reasoning in Multimodal Large Language Models}

While integrating Chain-of-Thought and Reinforcement Learning has significantly advanced LLMs~\cite{qwen2.5,yang2025qwen3}, video-domain adaptations~\cite{wang2025video,tang2025tspo,tian2025ego,chenscaling,wang2025videothinker} remain confined to offline batch-processing, incompatible with real-time streaming constraints. Conversely, strictly text-based streaming reasoning frameworks like StreamingThinker~\cite{tong2025streamingthinker} are ill-equipped for the concurrent think-and-respond demands and dense visual token bottlenecks of continuous video. Recognizing that reasoning advancements are fundamentally driven by RL, we introduce the first streaming RL approach to tackle these multimodal complexities. Furthermore, existing offline compression methods (e.g., LightThinker~\cite{zhang2025lightthinker}) merely reduce cache size without leveraging reasoning as active memory. In contrast, our work explicitly repurposes CoT trajectories as compressed semantic memory. By aggressively evicting dense visual tokens while preserving reasoning chains, we propose a mechanism that simultaneously scales reasoning capabilities and bounds KV cache growth for continuous video streaming.

\section{Streaming Watch–Think–Speak Paradigm}

\begin{figure*}[t]
\centering
\includegraphics[width=\linewidth]{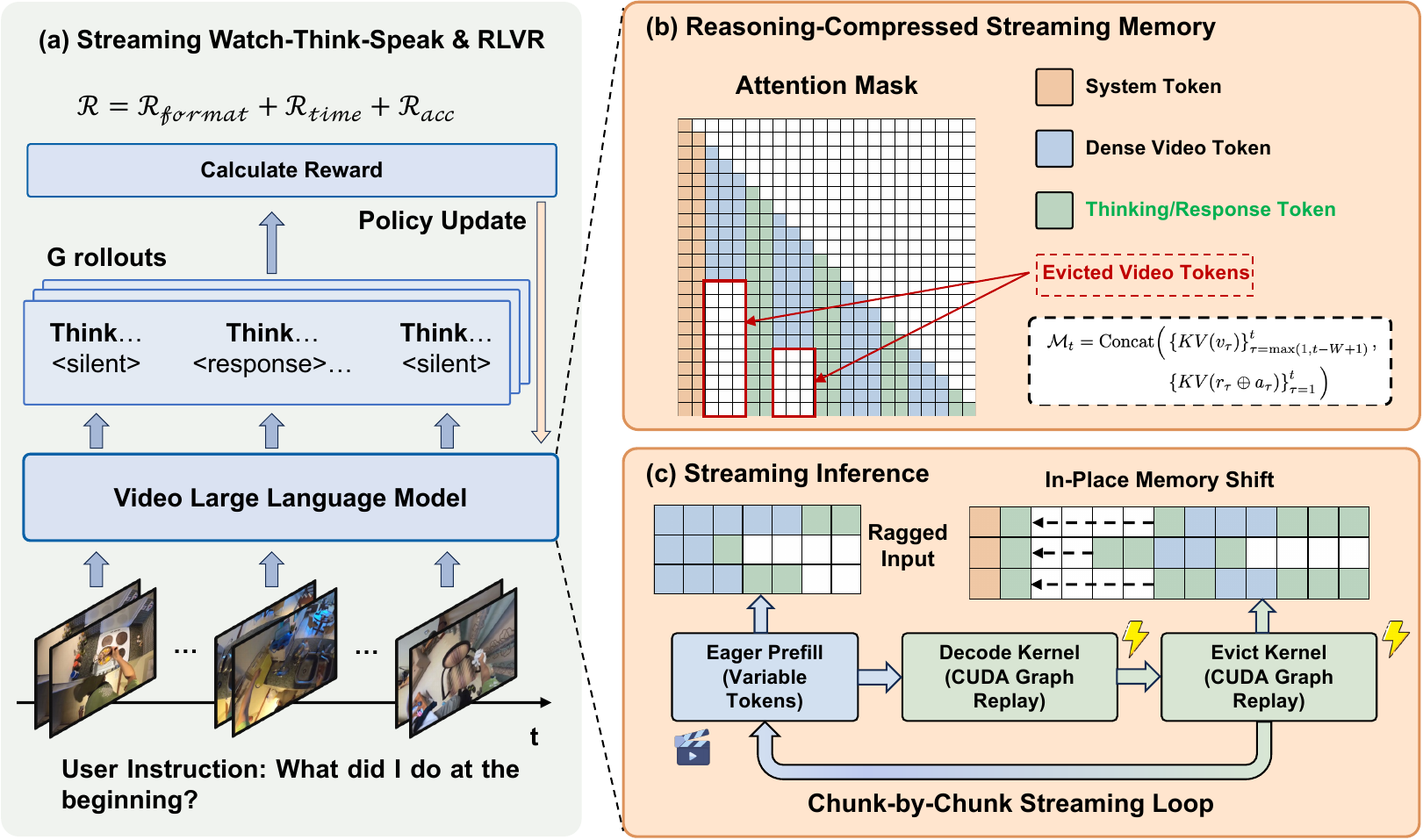}
\caption{Overview of the ThinkStream framework. (a) Streaming Watch-Think-Speak Paradigm \& RLVR: The model undergoes streaming rollouts and policy updates driven by format, latency, and accuracy rewards. (b) Reasoning-Compressed Streaming Memory: Outdated dense video tokens are dynamically evicted from the KV cache, while highly compressed reasoning and response tokens are retained as long-term semantic anchors. (c) Streaming Inference: A custom backend utilizes Eager Prefill for variable tokens and replayable CUDA Graphs for both the Decode and Evict Kernels, enabling an efficient chunk-by-chunk streaming loop with in-place memory shifting.}
\label{fig:method}
\end{figure*}

Human perception in continuous environments rarely proceeds as a purely reactive process. As new observations arrive, people constantly form provisional interpretations, refine them with additional evidence, and occasionally act when the accumulated understanding becomes sufficiently reliable. This process interleaves perception and reasoning: local observations trigger short reasoning updates, and these incremental updates gradually consolidate into a coherent global interpretation of the scene.

\subsection{Paradigm Design} 
Inspired by this cognitive pattern, we formulate streaming video understanding as a Watch–Think–Speak loop. Instead of separating perception and reasoning into two independent stages, the model performs incremental reasoning alongside the incoming stream. Each newly observed video segment triggers a short reasoning update that integrates the latest evidence with previously accumulated understanding. Over time, these updates progressively transform local perceptual signals into a structured interpretation of the evolving scene.

Within this paradigm, reasoning operates as a lightweight but continuous process. Upon observing a new video chunk, the model may (1) summarize newly observed events, (2) update hypotheses about ongoing activities or (3) refine causal or temporal relations inferred earlier. These reasoning steps are intentionally short and localized, ensuring that they can be generated within strict latency constraints while still allowing the model to maintain a coherent internal narrative of the stream.

Following each reasoning update, the model determines whether to respond to the user. When sufficient evidence has accumulated, the model produces an explicit answer; otherwise it remains silent and continues observing the stream.

\subsection{Formal Definition}
Let the incoming video stream be represented as a sequence of temporal chunks
$\mathcal{V} = \{v_1, v_2, \dots, v_t, \dots\}$,where $v_t$ denotes the video segment observed at step $t$. Let $\mathcal{I}$ denote the user instruction or dialogue context. The model maintains a historical state $\mathcal{H}_{t-1}$ summarizing the accumulated context from previous steps.

Operationally, each streaming step produces a structured output consisting of a reasoning segment followed by an interaction decision. The output format is defined as $\langle \texttt{think} \rangle ~ r_t ~ \langle/\texttt{think}\rangle 
\;a_t$.
Here $r_t$ denotes the reasoning tokens generated at step $t$, and $a_t$ represents the action taken by the model. The action either produces a response or explicitly indicates that the model remains silent.

Formally, given the incoming chunk $v_t$, the generation process follows

\begin{equation}
p(r_t, a_t \mid \mathcal{H}_{t-1}, v_t, \mathcal{I})
=
\prod_{i=1}^{|r_t \oplus a_t|}
\pi_\theta \big(y_i \mid y_{<i}, \mathcal{H}_{t-1}, v_t, \mathcal{I}\big),
\end{equation}
where $\pi_\theta$ denotes the policy model and $\oplus$ represents sequence concatenation.

The action space is restricted to two possibilities:
\begin{equation}
a_t \in
\big\{
\langle \text{silent} \rangle,
\;
\langle \text{response} \rangle \oplus c_t
\big\},
\end{equation}
where $c_t$ denotes the generated response content. Emitting $\langle \text{silent} \rangle$ indicates that the model continues observing the stream without responding, while emitting $\langle \text{response} \rangle$ signals that the model has accumulated sufficient evidence to produce an answer.

This formulation allows reasoning to unfold incrementally along the temporal stream. Each reasoning segment $r_t$ updates the model’s internal understanding based solely on observations available up to time $t$, ensuring strict streaming causality. Meanwhile, the speaking decision $a_t$ transforms the evolving reasoning state into an interaction policy that determines \emph{when} the model should respond and when it should continue watching.


\section{ThinkStream}
\label{sec:methodology}
Building upon our Watch-Think-Speak loop, we introduce \textbf{ThinkStream}, a unified framework that enables reasoning-capable multimodal models to operate under the strict computational constraints of continuous video streams. 
The central challenge lies in reconciling two competing requirements: maintaining coherent long-horizon understanding over the video stream while preserving real-time inference efficiency.


\subsection{Reasoning-Compressed Streaming Memory}
\label{sec:RCSM}
Continuous video streams naturally produce a large number of visual tokens. 
If all tokens are retained in the KV cache, both memory usage and attention complexity grow monotonically with time, eventually rendering real-time inference infeasible. 
However, not all past visual observations are equally important. 
Early visual details can often be replaced by higher-level semantic summaries once the model has formed a stable interpretation of the scene.

Motivated by this observation, we introduce Reasoning-Compressed Streaming Memory (\textbf{RCSM}). 
The core idea of RCSM is to treat reasoning tokens as compressed semantic representations of earlier visual observations. 
Instead of storing the entire history of dense visual tokens, the model gradually replaces outdated visual features with the reasoning traces generated during the streaming reasoning process.

\paragraph{\textbf{Reasoning-as-Compression State.}}
In the Watch-Think-Speak paradigm, each reasoning segment $r_t$ reflects the model's interpretation of the newly observed video chunk $v_t$ conditioned on prior context. 
These reasoning tokens encode semantic abstractions such as events, temporal relations, and causal hypotheses derived from the raw visual input.

We therefore interpret the reasoning state as a \textit{compression operator} that transforms dense perceptual evidence into compact symbolic representations. 
Formally, let $KV(\cdot)$ denote the cached key-value representations of tokens. 
Instead of preserving all visual tokens $\{v_1,\dots,v_t\}$, the model maintains a mixed memory state consisting of recent visual tokens and accumulated reasoning tokens.

The memory state at step $t$ is defined as
\begin{equation}
\mathcal{M}_t =
\text{Concat}
\Big(
\{ KV(v_\tau) \}_{\tau = \max(1, t-W+1)}^{t},
\{ KV(r_\tau \oplus a_\tau) \}_{\tau=1}^{t}
\Big),
\end{equation}
where $W$ denotes the size of the visual sliding window. 
In this design, reasoning tokens act as long-term semantic anchors that preserve the essential interpretation of earlier observations. When the stream length exceeds the window $W$, the earliest visual tokens are evicted from the cache. Let $v_{t-W}$ denote the oldest visual chunk within the window. 
When a new chunk $v_{t+1}$ arrives, the tokens corresponding to $v_{t-W}$ are removed from the KV cache. This design establishes a natural transition from dense perceptual memory to compressed semantic memory as the stream evolves.

Specifically, the number of retained visual tokens is upper-bounded by the window size, while reasoning tokens grow at a much slower rate due to their compact representation. Consequently, the effective context length remains stable even for long video streams. 
This property allows ThinkStream to sustain real-time inference over continuous video input while preserving the semantic information required for long-horizon reasoning.

\subsection{Streaming-Context RLVR Training}

RCSM specifies \emph{what} state is retained under streaming constraints; the remaining question is \emph{how} to train the model to produce reasoning traces that are simultaneously useful for solving the current query and informative enough to serve as compressed long-term memory once dense visual tokens are evicted.
To this end, we adopt a Streaming Reinforcement Learning with Verifiable Rewards (RLVR) setup tailored to the Watch-Think-Speak loop. 

\paragraph{\textbf{Streaming Rollout with Partial Reasoning.}}
Let the video stream be represented as a sequence of chunks $\mathcal{V} = \{v_1, v_2, \dots \}$ with user instruction $\mathcal{I}$. At each step $t$, the policy observes the current chunk $v_t$ alongside the accumulated streaming history $\mathcal{H}_{t-1}$ to produce an output tuple $o_t = (r_t, a_t)$. 

The streaming history, maintained via the RCSM memory mechanism in \cref{sec:RCSM}, is denoted as:
\begin{equation}
    \mathcal{H}_{t-1} = \{(v_\tau, r_\tau, a_\tau)\}_{\tau=1}^{t-1}
\end{equation}

Since the model only has access to the temporal prefix $\mathcal{V}_{\le t}$ at any given moment, the joint probability of the streaming rollout trajectory factorizes step-by-step as follows:
\begin{equation}
    \pi_\theta(o_{1:T} | \mathcal{V}_{1:T}, \mathcal{I}) = \prod_{t=1}^{T} \pi_\theta(r_t, a_t | \mathcal{H}_{t-1}, v_t, \mathcal{I})
\end{equation}



\paragraph{\textbf{RL Optimization.}} 
We adopt Group Relative Policy Optimization (GRPO) for model training. 
We sample $G$ trajectories 
$\{o^{(1)}_{1:T},\dots,o^{(G)}_{1:T}\}$ for each streaming instance, and optimize the policy using the
standard clipped objective with KL regularization:
\begin{equation}
\small
\mathcal{J}(\theta) =
\mathbb{E}
\left[
\frac{1}{G}\sum_{i=1}^{G}
\min
\left(
\rho_i(\theta)\hat{A}_i,
\text{clip}(\rho_i(\theta),1-\epsilon,1+\epsilon)\hat{A}_i
\right)
-
\beta D_{KL}(\pi_\theta\|\pi_{\text{ref}})
\right].
\end{equation}

\paragraph{\textbf{Reward Design.}}
The reward function provides the essential training signal to align the policy with the Watch-Think-Speak loop under stringent streaming constraints. To this end, we formulate a rule-based reward system comprising three integral components: an accuracy reward, a format reward, and a time reward. The \textit{accuracy reward} ($\mathcal{R}_{acc}$) assigns positive credit to responses that successfully match the ground-truth answer. To facilitate reliable automated verification, our RLVR samples are meticulously constructed utilizing deterministic formats, including multiple-choice, binary (Yes/No), and counting queries. Concurrently, the \textit{format reward} ($\mathcal{R}_{format}$) enforces strict adherence to the structured interaction protocol mandated by the loop, yielding a positive value strictly when the generated output conforms to the prescribed structural constraints, and zero otherwise.

Additionally, because streaming interactions necessitate temporally appropriate responses, we introduce a \textit{time reward} ($\mathcal{R}_{time}$). This component quantifies the temporal discrepancy between the model's response step $t_{\text{resp}}$ and the ground-truth step $t_{gt}$. We apply a linear decay within a predefined tolerance window $w$, formulated as $\mathcal{R}_{time} = \max(0, 1 - |t_{\text{resp}} - t_{gt}|/w)$. This mechanism effectively penalizes both premature inferences and excessive latency.

Ultimately, the total reward is computed as the linear summation of these three components:
\begin{equation}
\mathcal{R} = \mathcal{R}_{format} + \mathcal{R}_{time} + \mathcal{R}_{acc}.
\end{equation}
Collectively, this composite reward structure rigorously incentivizes the model to generate well-structured reasoning, actuate responses at optimal temporal junctures, and maintain factual fidelity throughout the duration of the streaming interaction.

\subsection{High-Efficiency Streaming Inference}

The online rollout phase of RLVR requires high-throughput streaming
inference, where the model repeatedly performs chunk-level decoding
while dynamically updating the KV cache. In particular, the RCSM memory
mechanism requires pruning stale visual tokens during decoding, which
demands explicit control over KV cache manipulation.

Existing frameworks are not well suited for this setting. Native
\texttt{transformers} implementations allow manual KV cache operations
but suffer from significant kernel launch overhead, while optimized
engines such as \texttt{vLLM} and \texttt{SGLang} achieve high decoding throughput but
provide limited support for customized KV cache updates required by RCSM.

To address this limitation, we design a streaming inference backend
based on \textit{CUDA Graphs}, which is illustrated in \cref{fig:method}(c). The execution pipeline consists of two
stages: \textbf{(1) Prefill Phase:}
Executed in eager mode to process newly arrived visual tokens and
update the KV cache. 
\textbf{(2) Decode-and-Prune Phase:}
Captured as a CUDA graph that performs token decoding together with KV cache pruning in replayable execution graphs. More implementation details can be found in the Appendix.

\section{ThinkStream Dataset}
\label{sec:data_construction}
To train the model for real-time streaming reasoning and interaction, we construct a large-scale, high-quality dataset featuring time-grounded CoT and verifiable question-answering pairs. Our data generation pipeline consists of three sequential stages.

\paragraph{\textbf{Video Segmentation and Dense Captioning.}} We employ PySceneDetect to segment continuous videos into coherent temporal scenes. We then utilize Qwen3-VL-235B-A22B-Instruct~\cite{bai2025qwen3} to densely caption each segment, yielding a sequence of timestamped semantic descriptions that serve as the foundational factual grounding for all subsequent generation phases.

\paragraph{\textbf{Diverse Instruction Synthesis.}} We construct highly diverse instruction data by exploring three key scenario dimensions: interaction modes (\textit{Real-time Dialogue}, \textit{Event Trigger}, \textit{Continuous Output}), temporal scopes (\textit{Past}, \textit{Current}, \textit{Future}), and seven fine-grained content semantics (including \textit{entity attributes}, \textit{action semantics}, \textit{spatial relationships}, \textit{causal reasoning}, \textit{procedural states}, \textit{global scenes}, and \textit{optical character recognition}). By computing the Cartesian product of these dimensions and filtering for practical validity, we derive \textit{39} meaningful scenario combinations. Guided by these filtered combinations, we synthesize diverse question-answering pairs that span multiple formats, specifically open-ended, multiple-choice, binary (Yes/No), and counting questions.

\paragraph{\textbf{Time-Grounded CoT Generation.}} Leveraging the timestamped captions and the generated instruction pairs, we simulate the internal reasoning traces to synthesize a dense, continuous, and strictly time-grounded CoT. This ensures that every thought and response is causally constrained to its specific timestamp, effectively acting as an uninterrupted \textit{stream of consciousness}.

Through this unified pipeline, we ultimately construct 110K Cold Start instances paired with detailed reasoning traces, alongside 9K RLVR instances strictly formatted for verifiable reward optimization. Comprehensive details regarding the video source, dataset distributions and prompt templates are provided in the Appendix.

\section{Experiments}
\label{sec:experiments}

\begin{table*}[t]
    \centering
    \caption{Performance comparison of ThinkStream and existing open-source offline and online models on the OVO-Bench streaming video benchmark. ThinkStream-3B achieves a strong average score, significantly surpassing its base model and competing online models.}
    \label{tab:ovo}
    \resizebox{\textwidth}{!}{
    \begin{tabular}{l | cccccc | c | ccc | c | c}
        \toprule
        \multirow{2}{*}{\textbf{Model}} & \multicolumn{7}{c|}{\textbf{Real-Time Visual Perception}} & \multicolumn{4}{c|}{\textbf{Backward Tracing}} & \multirow{2}{*}{\textbf{Avg.}} \\
        \cline{2-12}
        & OCR & ACR & ATR & STU & FPD & OJR & Avg. & EPM & ASI & HLD & Avg. & \\
        \midrule
        \multicolumn{13}{c}{\textbf{Open-source Offline Models}} \\
        \midrule
        LLaVA-Video-7B~\cite{zhang2024llava} & 69.13 & 58.72 & 68.83 & \textbf{49.44} & \textbf{74.26} & 59.78 & 63.52 & 56.23 & 57.43 & 7.53 & 40.4 & 51.96 \\
        Qwen2-VL-7B~\cite{wang2024qwen2} & 60.4 & 50.46 & 56.03 & 47.19 & 66.34 & 55.43 & 55.98 & 47.81 & 35.48 & 56.08 & 46.46 & 51.22 \\
        LongVU-7B~\cite{shen2024longvu} & 53.69 & 53.21 & 62.93 & 47.75 & 68.32 & 59.78 & 57.61 & 40.74 & 59.46 & 4.84 & 35.01 & 46.31 \\
        Qwen2.5-VL-3B~\cite{Qwen2.5-VL} & 76.51 & 44.03 & 67.24 & 42.13 & 68.31 & 61.96 & 60.03 & 50.50 & 53.38 & 22.04 & 41.98 & 51.00 \\
        Qwen2.5-VL-7B~\cite{Qwen2.5-VL} & 67.79 & 55.05 & 67.24 & 42.13 & 66.34 & 60.87 & 59.90 & 51.52 & 58.78 & 23.66 & 44.65 & 52.28 \\
        Qwen2.5-VL-32B~\cite{Qwen2.5-VL} & 77.18 & 58.72 & 68.10 & 50.56 & 74.26 & 57.61 & 64.40 & 58.59 & 62.84 & 29.57 & 50.33 & 57.37 \\
        \midrule
        \multicolumn{13}{c}{\textbf{Open-source Online Models}} \\
        \midrule
        Flash-VStream-7B~\cite{zhang2025flash} & 24.16 & 29.36 & 28.45 & 33.71 & 25.74 & 28.8 & 28.37 & 39.06 & 37.16 & 5.91 & 27.38 & 27.88 \\
        VideoLLM-online-8B~\cite{chen2024videollm} & 8.05 & 23.85 & 12.07 & 14.04 & 45.54 & 21.2 & 20.79 & 22.22 & 18.8 & 12.18 & 17.73 & 19.26 \\
        Dispider-7B~\cite{qian2025dispider} & 57.72 & 49.54 & 62.07 & 44.94 & 61.39 & 51.63 & 54.55 & 48.48 & 55.41 & 4.3 & 36.06 & 45.31 \\
        StreamForest-7B~\cite{zeng2025streamforest} & 68.46 & 53.21 & \textbf{71.55} & 47.75 & 65.35 & 60.87 & 61.20 & \textbf{58.92} & \textbf{64.86} & 32.26 & 52.02 & 56.61 \\
        Streamo-3B~\cite{xia2025streaming} & 78.52 & 52.29 & 67.24 & 44.38 & 55.45 & \textbf{71.20} & 61.51 & 51.18 & 57.43 & 16.67 & 41.76 & 51.64 \\
        \tableLineColor \textbf{ThinkStream-3B (Ours)} & \textbf{85.23} & \textbf{64.22} & \underline{69.82} & \textbf{49.43} & \textbf{69.31} & \underline{64.13} & \textbf{67.03} & \underline{53.87} & \underline{59.46} & \textbf{43.55} & \textbf{52.30} & \textbf{59.66} \\
        \bottomrule
    \end{tabular}
    }
\end{table*}

To evaluate the effectiveness of the proposed ThinkStream framework, we conduct extensive experiments across streaming and offline video benchmarks. We also provide detailed ablation studies and efficiency profiling to validate our architectural designs and the real-time capabilities of our custom streaming inference engine.

\paragraph{\textbf{Implementation Details.}} We initialize our framework based on the Qwen2.5-VL-3B~\cite{Qwen2.5-VL} model. During the Cold Start phase, we train the model with a batch size of 64 and a learning rate of $1 \times 10^{-5}$. In the subsequent Reinforcement Learning (RL) phase, we employ a batch size of 8, a GRPO group size of 8, and a learning rate of $2 \times 10^{-7}$. The AdamW~\cite{loshchilov2017decoupled} optimizer is utilized across all training stages. For video processing, frames are sampled at a rate of 2 FPS and encoded using native dynamic resolution. All training experiments are conducted on a single node equipped with 8 NVIDIA H20 GPUs.

\begin{table*}[t]
    \centering
    \caption{Performance comparison of various MLLMs on the StreamingBench Real-Time benchmark. ThinkStream-3B demonstrates highly competitive performance against proprietary models such as GPT-4o and vastly exceeds other open-source online MLLMs.}
    \label{tab:streaming}
    \resizebox{\textwidth}{!}{
    \begin{tabular}{l | cccccccccc | c}
        \toprule
        \textbf{Model} & \textbf{OP} & \textbf{CR} & \textbf{CS} & \textbf{ATP} & \textbf{EU} & \textbf{TR} & \textbf{PR} & \textbf{SU} & \textbf{ACP} & \textbf{CT} & \textbf{Avg.} \\
        \midrule
        Human & 89.47 & 92.00 & 93.60 & 91.47 & 95.65 & 92.52 & 88.00 & 88.75 & 89.74 & 91.30 & 91.46 \\
        \midrule
        \multicolumn{12}{c}{\textbf{Proprietary MLLMs}} \\
        \midrule
        Gemini 1.5 pro~\cite{gemini20241} & 79.02 & 80.47 & 83.54 & 79.67 & 80.00 & 84.74 & 77.78 & 64.23 & 71.95 & 48.70 & 75.69 \\
        GPT-4o~\cite{hurst2024gpt} & 77.11 & 80.47 & 83.91 & 76.47 & 70.19 & 83.80 & 66.67 & 62.19 & 69.12 & 49.22 & 73.28 \\
        Claude 3.5 Sonnet & 73.33 & 80.47 & 84.09 & 82.02 & 75.39 & 79.53 & 61.11 & 61.79 & 69.32 & 43.09 & 72.44 \\
        \midrule
        \multicolumn{12}{c}{\textbf{Open-source Offline MLLMs}} \\
        \midrule
        VILA-1.5-8B~\cite{lin2024vila} & 53.68 & 49.22 & 70.98 & 56.86 & 53.42 & 53.89 & 54.63 & 48.78 & 50.14 & 17.62 & 52.32 \\
        LongVA-7B~\cite{zhang2024long} & 70.03 & 63.28 & 61.20 & 70.92 & 62.73 & 59.50 & 61.11 & 53.66 & 54.67 & 34.72 & 59.96 \\
        LLaVA-NeXT-Video-32B~\cite{liu2024llavanext} & 78.20 & 70.31 & 73.82 & 76.80 & 63.35 & 69.78 & 57.41 & 56.10 & 64.31 & 38.86 & 66.96 \\
        Qwen2.5-VL-3B~\cite{Qwen2.5-VL} & 67.21 & 68.75 & 75.39 & 79.17 & 72.96 & 72.27 & 71.30 & 61.38 & 71.59 & 26.06 & 67.96 \\
        Qwen2.5-VL-7B~\cite{Qwen2.5-VL} & 77.93 & 76.56 & 78.55 & 80.86 & 76.73 & 76.95 & 80.56 & 65.45 & 65.72 & 52.85 & 73.31 \\
        Qwen2.5-VL-32B~\cite{Qwen2.5-VL} & 76.29 & 79.69 & 78.55 & 83.50 & 76.10 & 79.44 & 80.56 & 61.38 & 68.27 & 59.07 & 74.27  \\
        \midrule
        \multicolumn{12}{c}{\textbf{Open-source Online MLLMs}} \\
        \midrule
        Flash-VStream-7B~\cite{zhang2025flash} & 25.89 & 43.57 & 24.91 & 23.87 & 27.33 & 13.08 & 18.52 & 25.20 & 23.87 & 48.70 & 23.23 \\
        VideoLLM-online-8B~\cite{chen2024videollm} & 39.07 & 40.06 & 34.49 & 31.05 & 45.96 & 32.40 & 31.48 & 34.16 & 42.49 & 27.89 & 35.99 \\
        Dispider-7B~\cite{qian2025dispider} & 74.92 & \textbf{75.53} & 74.10 & 73.08 & 74.44 & 59.92 & 76.14 & 62.91 & 62.16 & \textbf{45.80} & 67.63 \\
        \tableLineColor \textbf{ThinkStream-3B (Ours)} & \textbf{83.74} & \underline{70.31} & \textbf{76.03} & \textbf{82.69} & \textbf{76.73} & \textbf{78.19} & \textbf{76.85} & \textbf{70.73} & \textbf{74.43} & \underline{45.21} & \textbf{75.00} \\
        \bottomrule
    \end{tabular}
    }
\end{table*}

\subsection{Main Results}

\paragraph{\textbf{Streaming Video Benchmarks.}} We comprehensively evaluate ThinkStream on specialized streaming video benchmarks~\cite{niu2025ovo,lin2024streamingbench}. As shown in ~\cref{tab:ovo} and ~\cref{tab:streaming}, our models consistently outperform existing models. Specifically, in ~\cref{tab:ovo}, ThinkStream-3B achieves an overall average score of 59.66, significantly surpassing both its base model Qwen2.5-VL-3B (51.00) and competing open-source online models such as Streamo-3B (51.64) on OVO-Bench. Furthermore, on the StreamingBench Real-Time detailed in ~\cref{tab:streaming}, ThinkStream-3B attains an average score of 75.00. This not only vastly exceeds other open-source online MLLMs like Dispider-7B (67.63) but also demonstrates highly competitive performance against proprietary models such as GPT-4o (73.28). The results demonstrate that ThinkStream achieves strong performance on streaming video tasks, showing the effectiveness of our Watch-Think-Speak loop.

\paragraph{\textbf{Offline Video Benchmarks.}} To ensure the general ability of our streaming-optimized architecture, we also evaluate on standard offline video benchmarks~\cite{fu2025video,wu2024longvideobench}. Despite aggressively evicting visual tokens, ThinkStream-3B achieves highly competitive performance as demonstrated in ~\cref{tab:offline}. Specifically, our model achieves a score of 61.9 on VideoMME and 56.4 on Long VideoBench. With an overall average score of 59.4 across the evaluated metrics, ThinkStream-3B significantly outperforms its base model Qwen2.5-VL-3B (with a 54.4 average). This proves that our ThinkStream effectively preserves understanding capabilities on offline video tasks.

\begin{table}[t]
\centering
\caption{Model performance comparison on standard offline video benchmarks, including VideoMME and Long VideoBench. Despite aggressively evicting visual tokens, ThinkStream-3B achieves highly competitive performance, demonstrating that it effectively preserves understanding capabilities on offline tasks.}
\label{tab:offline}
\resizebox{\textwidth}{!}{
\begin{tabular}{lccccc}
\toprule
\textbf{Model} & \begin{tabular}[c]{@{}c@{}}\textbf{OVO} \\ \textbf{Real-Time}\end{tabular} & \begin{tabular}[c]{@{}c@{}}\textbf{OVO} \\ \textbf{Backward}\end{tabular} & \textbf{VideoMME} & \textbf{LongVideoBench} & \textbf{Avg.} \\
\midrule
Flash-VStream-7B & 28.4 & 27.4 & 61.2 & - & - \\
VideoLLM-online-8B & 20.8 & 17.7 & 26.9 & - & - \\
Dispider-7B & 54.6 & 36.1 & 57.2 & - & - \\
Streamo-3B & 61.5 & 41.8 & 61.8 & 56.2 & 55.3 \\
Qwen2.5-VL-3B & 60.0 & 41.9 & 61.5 & 54.2 & 54.4 \\
\tableLineColor ThinkStream-3B (Ours) & \textbf{67.0 (\increasingData{7.0})} & \textbf{52.3 (\increasingData{10.4})} & \textbf{61.9 (\increasingData{0.8})} & \textbf{56.4 (\increasingData{1.8})} & \textbf{59.4 (\increasingData{5.0})} \\
\bottomrule
\end{tabular}
}
\end{table}

\subsection{Ablation Study}

\paragraph{\textbf{Reasoning Token Budget.}} We ablate the maximum reasoning token budget allocated per second of video. As illustrated in ~\cref{tab:ablation_token_budget}, increasing the reasoning token budget up to 20 tokens per second yields substantial performance improvements, as the model gains sufficient capacity for logical deduction. Specifically, scaling the budget from 0 to 20 tokens significantly improves the OVO-Backward score from 41.8 to 52.3. However, allocating beyond 20 tokens per second results in marginal performance gains (the OVO-Backward score only reaches 52.6 at 30 tokens) while inflating computational overhead and decoding latency, which jumps from 380 ms at 20 tokens to 505 ms at 30 tokens. Thus, 20 tokens per second strikes the optimal balance between reasoning capacity and efficiency.

\begin{table}[t]
    \centering
    \begin{minipage}[t]{0.48\textwidth}
        \centering
        \caption{Ablation on Reasoning Token Budget. Allocating 20 tokens per second strikes the optimal balance between reasoning capacity and efficiency.}
        \resizebox{\linewidth}{!}{
        \begin{tabular}{l | c c c}
            \toprule
            \textbf{\#Token} & \textbf{Streaming.} & \textbf{OVO-BW} & \textbf{Latency (ms) $\downarrow$} \\
            \midrule
            0 & 69.6 & 41.8 & \textbf{130} \\
            5 & 70.2 & 46.9 & 193 \\
            10 & 72.3 & 49.7 & 255 \\
            \tableLineColor 20 & \textbf{75.0} & \underline{52.3} & 380 \\
            30 & \textbf{75.0} & \textbf{52.6} & 505 \\
            \bottomrule
        \end{tabular}
        }
        \label{tab:ablation_token_budget}
    \end{minipage}\hfill
    \begin{minipage}[t]{0.48\textwidth}
        \centering
        \caption{Ablation on Visual KV Cache Window Size. Setting the window size to 20 seconds achieves the best trade-off between accuracy and efficiency.}
        \resizebox{\linewidth}{!}{
        \begin{tabular}{l | c c c}
            \toprule
            \textbf{Window} & \textbf{Streaming.} & \textbf{OVO-RT} & \textbf{OVO-BW} \\
            \midrule
            5s & 73.9 & 65.2 & 46.7 \\
            10s & 74.5 & 66.3 & 50.1 \\
            \tableLineColor 20s & \textbf{75.0} & \textbf{67.0} & \textbf{52.3} \\
            30s & 74.9 & 66.8 & 51.6 \\
            \bottomrule
        \end{tabular}
        }
        \label{tab:ablation_window_size}
    \end{minipage}
\end{table}

\begin{table}[t]
    \small
    \centering
    \caption{Ablation on Stream Memory Representation and RLVR Optimization. Post-training with RLVR demonstrates that reasoning tokens act as a highly compressed long-term memory representation.}
    \begin{tabular}{l | c c c c}
        \toprule
        \textbf{Memory Variant} & \textbf{Streaming.} & \textbf{OVO-BW} & \textbf{OVO-RT} & \textbf{Avg.} \\
        \midrule
        Qwen2.5-VL-3B & 68.9 & 41.9 & 60.0 & 56.9 \\
        Discrete caption as memory & 59.0 & 36.9 & 50.1 & 48.7 \\
        No memory & 69.6 & 41.8 & 59.2 & 56.9 \\
        Cold-start CoT memory & 70.6 & 47.6 & 63.3 & 60.5 \\
        \tableLineColor RLVR-optimized CoT memory & \textbf{75.0} & \textbf{52.3} & \textbf{67.0} & \textbf{64.8} \\
        \bottomrule
    \end{tabular}
    \label{tab:ablation_memory_rlvr}
\end{table}

\paragraph{\textbf{Video KV Cache Window Size.}} We investigate the impact of the dense visual token retention window, which is shown in ~\cref{tab:ablation_window_size}. Setting the window size to 20 seconds achieves the best performance, peaking at 75.0 on StreamingBench Real-Time and 52.3 on OVO-Backward. By contrast, narrower windows such as 5s and 10s result in lower OVO-Backward scores of 46.7 and 50.1, respectively, whereas expanding the window to 30s slightly degrades the OVO-Backward performance to 51.6. This confirms the validity of our assumption that short-term, high-resolution visual context combined with long-term semantic reasoning tokens is sufficient for continuous video understanding.

\paragraph{\textbf{Stream Memory Representation and RLVR Optimization.}} To validate the necessity of our RLVR pipeline and the specific memory representation, we compare several variants: (1) no memory, (2) discrete caption tokens as memory, (3) Cold-start CoT memory, and (4) our RLVR-optimized CoT memory. As is shown in ~\cref{tab:ablation_memory_rlvr}, initializing the model with constructed cold-start CoT data yields a significant gain, improving the average score from 56.9 (no memory) to 60.5. Interestingly, utilizing discrete caption tokens actually diminishes the overall capability, dropping the average score to 48.7. Finally, ost-training the model with RLVR further boosts average performance by 4.3 points (reaching 64.8), demonstrating that reasoning tokens learn to act as a far superior, highly compressed long-term memory compared to naive discrete captions.

\begin{figure}[t]
    \centering
    \begin{minipage}{0.48\textwidth}
        \centering
        \includegraphics[width=\linewidth]{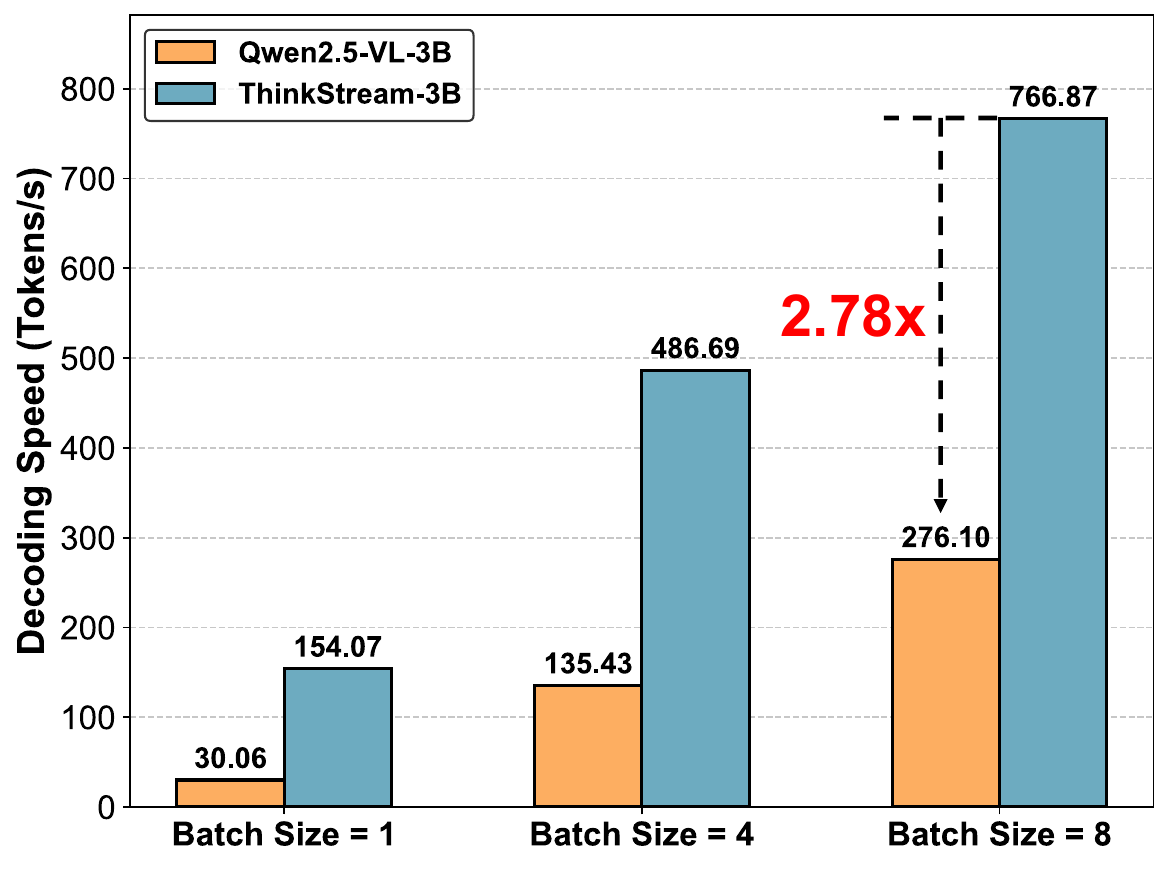}
        \caption{Token decoding speed comparison across different batch sizes. The custom CUDA Graph-based streaming inference engine achieves a massive speedup compared to the standard Qwen2.5-VL-3B baseline, maintaining high throughput while preserving flexible KV cache control.}
        \label{fig:efficiency_speed}
    \end{minipage}
    \hfill
    \begin{minipage}{0.48\textwidth}
        \centering
        \includegraphics[width=\linewidth]{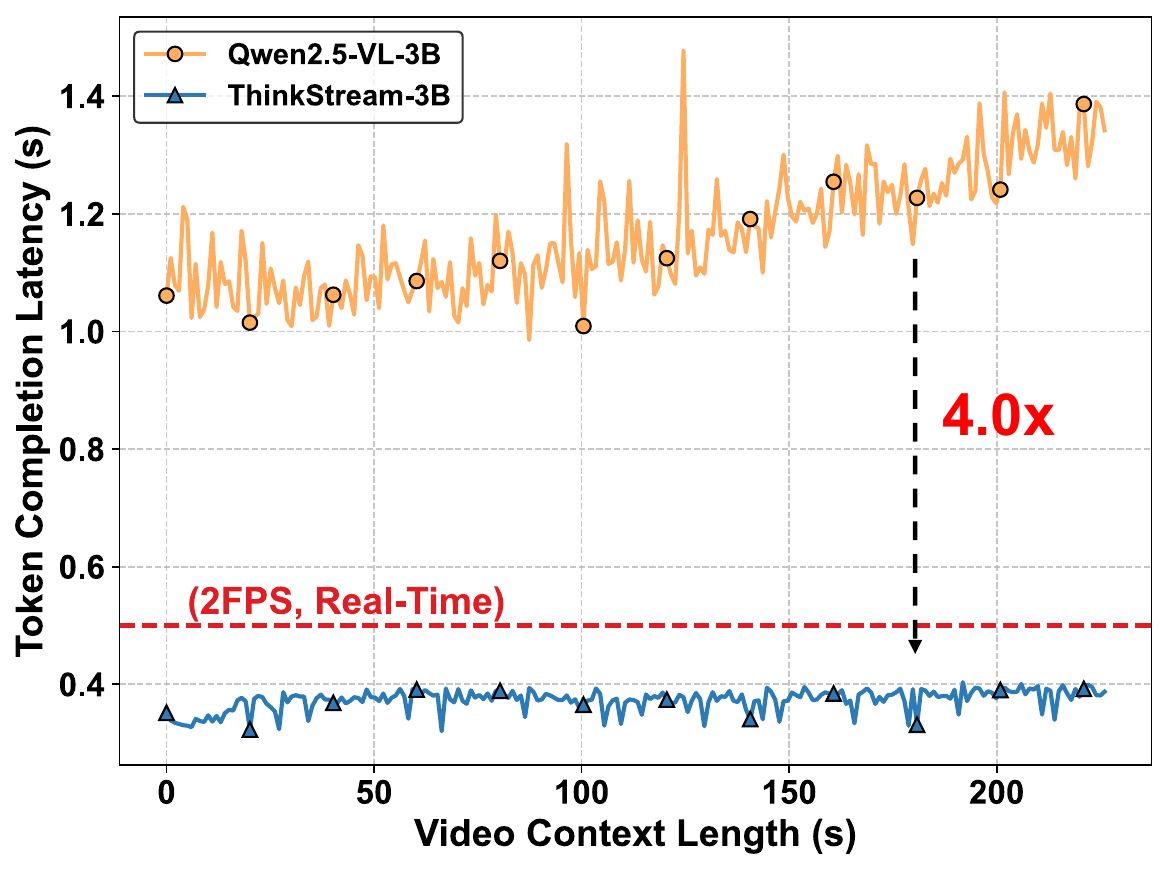}
        \caption{Real-time latency scaling with processed video length. ThinkStream successfully bounds the end-to-end inference latency below the 0.5s real-time threshold (required for 2 FPS inputs) as the video context grows, whereas the baseline model scales poorly and consistently violates the threshold.}
        \label{fig:efficiency_latency}
    \end{minipage}
\end{figure}

\subsection{Efficiency and Real-Time Analysis}

A core contribution of our work is ensuring that the Streaming Watch-Think-Speak paradigm strictly adheres to real-time constraints. We profile our CUDA Graph-based Streaming Inference Engine against standard eager transformers implementations.

As shown in \cref{fig:efficiency_speed}, our custom engine achieves a massive speedup in token decoding across various batch sizes compared to the Qwen2.5-VL-3B model. For instance, at a batch size of 1, our engine delivers 154.07 tokens/s compared to the baseline's 30.06 tokens/s, representing a more than 5$\times$ speedup. Furthermore, at a batch size of 8, our decoding speed scales efficiently to 766.87 tokens/s, significantly outperforming the baseline's 276.10 tokens/s. This confirms that our custom backend maintains high throughput while preserving the flexible, manual KV cache control required for our eviction strategy.

Crucially, as illustrated in \cref{fig:efficiency_latency}, our framework successfully bounds latency as the processed video length increases. While the baseline model's latency scales poorly and consistently violates the real-time threshold—fluctuating between 1.0s and 1.4s per second of video—our combination of algorithmic KV cache eviction and engineering optimizations ensures that the end-to-end inference latency remains flat. It consistently stays below the 0.5s real-time threshold required for 2 FPS inputs. The total processing delay remains bounded, proving that our approach can be deployed efficiently in streaming video environments.

\section{Conclusion}
We study streaming video reasoning, where models must continuously interpret incoming observations under strict latency and memory constraints. 
We introduce the Watch--Think--Speak paradigm and propose ThinkStream, a framework that enables incremental reasoning over long video streams. 
Through Reasoning-Compressed Streaming Memory (RCSM) and streaming reinforcement learning with verifiable rewards, our approach maintains bounded memory usage while preserving coherent long-horizon understanding. 
Experiments show that ThinkStream achieves strong performance on streaming video benchmarks with low latency while retaining competitive capability on standard offline video tasks.



%
%
\bibliographystyle{splncs04}
\bibliography{main}

@article{wang2024qwen2,
  title={Qwen2-vl: Enhancing vision-language model's perception of the world at any resolution},
  author={Wang, Peng and Bai, Shuai and Tan, Sinan and Wang, Shijie and Fan, Zhihao and Bai, Jinze and Chen, Keqin and Liu, Xuejing and Wang, Jialin and Ge, Wenbin and others},
  journal={arXiv preprint arXiv:2409.12191},
  year={2024}
}

@article{bai2025qwen3,
  title={Qwen3-vl technical report},
  author={Bai, Shuai and Cai, Yuxuan and Chen, Ruizhe and Chen, Keqin and Chen, Xionghui and Cheng, Zesen and Deng, Lianghao and Ding, Wei and Gao, Chang and Ge, Chunjiang and others},
  journal={arXiv preprint arXiv:2511.21631},
  year={2025}
}

@article{Qwen2.5-VL,
  title={Qwen2.5-VL Technical Report},
  author={Bai, Shuai and Chen, Keqin and Liu, Xuejing and Wang, Jialin and Ge, Wenbin and Song, Sibo and Dang, Kai and Wang, Peng and Wang, Shijie and Tang, Jun and Zhong, Humen and Zhu, Yuanzhi and Yang, Mingkun and Li, Zhaohai and Wan, Jianqiang and Wang, Pengfei and Ding, Wei and Fu, Zheren and Xu, Yiheng and Ye, Jiabo and Zhang, Xi and Xie, Tianbao and Cheng, Zesen and Zhang, Hang and Yang, Zhibo and Xu, Haiyang and Lin, Junyang},
  journal={arXiv preprint arXiv:2502.13923},
  year={2025}
}

@article{loshchilov2017decoupled,
  title={Decoupled weight decay regularization},
  author={Loshchilov, Ilya and Hutter, Frank},
  journal={arXiv preprint arXiv:1711.05101},
  year={2017}
}

@article{lin2024streamingbench,
  title={Streamingbench: Assessing the gap for mllms to achieve streaming video understanding},
  author={Lin, Junming and Fang, Zheng and Chen, Chi and Wan, Zihao and Luo, Fuwen and Li, Peng and Liu, Yang and Sun, Maosong},
  journal={arXiv preprint arXiv:2411.03628},
  year={2024}
}

@inproceedings{niu2025ovo,
  title={Ovo-bench: How far is your video-llms from real-world online video understanding?},
  author={Niu, Junbo and Li, Yifei and Miao, Ziyang and Ge, Chunjiang and Zhou, Yuanhang and He, Qihao and Dong, Xiaoyi and Duan, Haodong and Ding, Shuangrui and Qian, Rui and others},
  booktitle={Proceedings of the Computer Vision and Pattern Recognition Conference},
  pages={18902--18913},
  year={2025}
}

@inproceedings{fu2025video,
  title={Video-mme: The first-ever comprehensive evaluation benchmark of multi-modal llms in video analysis},
  author={Fu, Chaoyou and Dai, Yuhan and Luo, Yongdong and Li, Lei and Ren, Shuhuai and Zhang, Renrui and Wang, Zihan and Zhou, Chenyu and Shen, Yunhang and Zhang, Mengdan and others},
  booktitle={Proceedings of the IEEE/CVF conference on computer vision and pattern recognition},
  pages={24108--24118},
  year={2025}
}

@article{wu2024longvideobench,
  title={Longvideobench: A benchmark for long-context interleaved video-language understanding},
  author={Wu, Haoning and Li, Dongxu and Chen, Bei and Li, Junnan},
  journal={Advances in Neural Information Processing Systems},
  volume={37},
  pages={28828--28857},
  year={2024}
}

@inproceedings{chen2024videollm,
  title={Videollm-online: Online video large language model for streaming video},
  author={Chen, Joya and Lv, Zhaoyang and Wu, Shiwei and Lin, Kevin Qinghong and Song, Chenan and Gao, Difei and Liu, Jia-Wei and Gao, Ziteng and Mao, Dongxing and Shou, Mike Zheng},
  booktitle={Proceedings of the IEEE/CVF Conference on Computer Vision and Pattern Recognition},
  pages={18407--18418},
  year={2024}
}

@inproceedings{zhang2025flash,
  title={Flash-vstream: Efficient real-time understanding for long video streams},
  author={Zhang, Haoji and Wang, Yiqin and Tang, Yansong and Liu, Yong and Feng, Jiashi and Jin, Xiaojie},
  booktitle={Proceedings of the IEEE/CVF international conference on computer vision},
  pages={21059--21069},
  year={2025}
}

@article{yang2025streammem,
  title={Streammem: Query-agnostic kv cache memory for streaming video understanding},
  author={Yang, Yanlai and Zhao, Zhuokai and Shukla, Satya Narayan and Singh, Aashu and Mishra, Shlok Kumar and Zhang, Lizhu and Ren, Mengye},
  journal={arXiv preprint arXiv:2508.15717},
  year={2025}
}

@inproceedings{kim2025egospeak,
  title={Egospeak: learning when to speak for egocentric conversational agents in the wild},
  author={Kim, Junhyeok and Kim, Min-Soo and Chung, Jiwan and Cho, Jungbin and Kim, Jisoo and Kim, Sungwoong and Sim, Gyeongbo and Yu, Youngjae},
  booktitle={Findings of the Association for Computational Linguistics: NAACL 2025},
  pages={2990--3005},
  year={2025}
}

@inproceedings{ding2025streammind,
  title={Streammind: Unlocking full frame rate streaming video dialogue through event-gated cognition},
  author={Ding, Xin and Wu, Hao and Yang, Yifan and Jiang, Shiqi and Zhang, Qianxi and Bai, Donglin and Chen, Zhibo and Cao, Ting},
  booktitle={Proceedings of the IEEE/CVF International Conference on Computer Vision},
  pages={13448--13459},
  year={2025}
}

@inproceedings{wang2025video,
  title={Video-rts: Rethinking reinforcement learning and test-time scaling for efficient and enhanced video reasoning},
  author={Wang, Ziyang and Yoon, Jaehong and Yu, Shoubin and Islam, Md Mohaiminul and Bertasius, Gedas and Bansal, Mohit},
  booktitle={Proceedings of the 2025 Conference on Empirical Methods in Natural Language Processing},
  pages={28114--28128},
  year={2025}
}

@article{tang2025tspo,
  title={TSPO: Temporal Sampling Policy Optimization for Long-form Video Language Understanding},
  author={Tang, Canhui and Han, Zifan and Sun, Hongbo and Zhou, Sanping and Zhang, Xuchong and Wei, Xin and Yuan, Ye and Xu, Jinglin and Sun, Hao},
  journal={arXiv preprint arXiv:2508.04369},
  year={2025}
}

@article{yang2025qwen3,
  title={Qwen3 technical report},
  author={Yang, An and Li, Anfeng and Yang, Baosong and Zhang, Beichen and Hui, Binyuan and Zheng, Bo and Yu, Bowen and Gao, Chang and Huang, Chengen and Lv, Chenxu and others},
  journal={arXiv preprint arXiv:2505.09388},
  year={2025}
}

@article{qwen2.5,
    title   = {Qwen2.5 Technical Report}, 
    author  = {An Yang and Baosong Yang and Beichen Zhang and Binyuan Hui and Bo Zheng and Bowen Yu and Chengyuan Li and Dayiheng Liu and Fei Huang and Haoran Wei and Huan Lin and Jian Yang and Jianhong Tu and Jianwei Zhang and Jianxin Yang and Jiaxi Yang and Jingren Zhou and Junyang Lin and Kai Dang and Keming Lu and Keqin Bao and Kexin Yang and Le Yu and Mei Li and Mingfeng Xue and Pei Zhang and Qin Zhu and Rui Men and Runji Lin and Tianhao Li and Tingyu Xia and Xingzhang Ren and Xuancheng Ren and Yang Fan and Yang Su and Yichang Zhang and Yu Wan and Yuqiong Liu and Zeyu Cui and Zhenru Zhang and Zihan Qiu},
    journal = {arXiv preprint arXiv:2412.15115},
    year    = {2024}
}

@article{tong2025streamingthinker,
  title={StreamingThinker: Large Language Models Can Think While Reading},
  author={Tong, Junlong and Fan, Yingqi and Zhao, Anhao and Ma, Yunpu and Shen, Xiaoyu},
  journal={arXiv preprint arXiv:2510.17238},
  year={2025}
}

@inproceedings{zhang2025lightthinker,
  title={Lightthinker: Thinking step-by-step compression},
  author={Zhang, Jintian and Zhu, Yuqi and Sun, Mengshu and Luo, Yujie and Qiao, Shuofei and Du, Lun and Zheng, Da and Chen, Huajun and Zhang, Ningyu},
  booktitle={Proceedings of the 2025 Conference on Empirical Methods in Natural Language Processing},
  pages={13318--13339},
  year={2025}
}

@article{zhang2024llava,
  title={Llava-video: Video instruction tuning with synthetic data},
  author={Zhang, Yuanhan and Wu, Jinming and Li, Wei and Li, Bo and Ma, Zejun and Liu, Ziwei and Li, Chunyuan},
  journal={arXiv preprint arXiv:2410.02713},
  year={2024}
}

@inproceedings{li2025breaking,
  title={Breaking the encoder barrier for seamless video-language understanding},
  author={Li, Handong and Zhang, Yiyuan and Guo, Longteng and Yue, Xiangyu and Liu, Jing},
  booktitle={Proceedings of the IEEE/CVF International Conference on Computer Vision},
  pages={23167--23176},
  year={2025}
}

@article{wang2025streambridge,
  title={Streambridge: Turning your offline video large language model into a proactive streaming assistant},
  author={Wang, Haibo and Feng, Bo and Lai, Zhengfeng and Xu, Mingze and Li, Shiyu and Ge, Weifeng and Dehghan, Afshin and Cao, Meng and Huang, Ping},
  journal={arXiv preprint arXiv:2505.05467},
  year={2025}
}

@article{xu2025streamingvlm,
  title={Streamingvlm: Real-time understanding for infinite video streams},
  author={Xu, Ruyi and Xiao, Guangxuan and Chen, Yukang and He, Liuning and Peng, Kelly and Lu, Yao and Han, Song},
  journal={arXiv preprint arXiv:2510.09608},
  year={2025}
}

@article{feng2025video,
  title={Video-r1: Reinforcing video reasoning in mllms},
  author={Feng, Kaituo and Gong, Kaixiong and Li, Bohao and Guo, Zonghao and Wang, Yibing and Peng, Tianshuo and Wu, Junfei and Zhang, Xiaoying and Wang, Benyou and Yue, Xiangyu},
  journal={arXiv preprint arXiv:2503.21776},
  year={2025}
}

@article{wu2025tempr1,
  title={TempR1: Improving Temporal Understanding of MLLMs via Temporal-Aware Multi-Task Reinforcement Learning},
  author={Wu, Tao and Yang, Li and Zhan, Gen and Zhang, Yabin and Liao, Yiting and Li, Junlin and Fu, Deliang and Zhang, Li and Wang, Limin},
  journal={arXiv preprint arXiv:2512.03963},
  year={2025}
}

@article{xia2025streaming,
  title={Streaming Video Instruction Tuning},
  author={Xia, Jiaer and Chen, Peixian and Zhang, Mengdan and Sun, Xing and Zhou, Kaiyang},
  journal={arXiv preprint arXiv:2512.21334},
  year={2025}
}

@article{shen2024longvu,
  title={Longvu: Spatiotemporal adaptive compression for long video-language understanding},
  author={Shen, Xiaoqian and Xiong, Yunyang and Zhao, Changsheng and Wu, Lemeng and Chen, Jun and Zhu, Chenchen and Liu, Zechun and Xiao, Fanyi and Varadarajan, Balakrishnan and Bordes, Florian and others},
  journal={arXiv preprint arXiv:2410.17434},
  year={2024}
}

@inproceedings{qian2025dispider,
  title={Dispider: Enabling video llms with active real-time interaction via disentangled perception, decision, and reaction},
  author={Qian, Rui and Ding, Shuangrui and Dong, Xiaoyi and Zhang, Pan and Zang, Yuhang and Cao, Yuhang and Lin, Dahua and Wang, Jiaqi},
  booktitle={Proceedings of the Computer Vision and Pattern Recognition Conference},
  pages={24045--24055},
  year={2025}
}

@article{zeng2025streamforest,
  title={Streamforest: Efficient online video understanding with persistent event memory},
  author={Zeng, Xiangyu and Qiu, Kefan and Zhang, Qingyu and Li, Xinhao and Wang, Jing and Li, Jiaxin and Yan, Ziang and Tian, Kun and Tian, Meng and Zhao, Xinhai and others},
  journal={arXiv preprint arXiv:2509.24871},
  year={2025}
}

@article{gemini20241,
  title={1.5: unlocking multimodal understanding across millions of tokens of context},
  author={Gemini, Gemini Team},
  journal={arXiv preprint arXiv:2403.05530},
  year={2024}
}

@article{hurst2024gpt,
  title={Gpt-4o system card},
  author={Hurst, Aaron and Lerer, Adam and Goucher, Adam P and Perelman, Adam and Ramesh, Aditya and Clark, Aidan and Ostrow, AJ and Welihinda, Akila and Hayes, Alan and Radford, Alec and others},
  journal={arXiv preprint arXiv:2410.21276},
  year={2024}
}

@inproceedings{lin2024vila,
  title={Vila: On pre-training for visual language models},
  author={Lin, Ji and Yin, Hongxu and Ping, Wei and Molchanov, Pavlo and Shoeybi, Mohammad and Han, Song},
  booktitle={Proceedings of the IEEE/CVF conference on computer vision and pattern recognition},
  pages={26689--26699},
  year={2024}
}

@article{zhang2024long,
  title={Long context transfer from language to vision},
  author={Zhang, Peiyuan and Zhang, Kaichen and Li, Bo and Zeng, Guangtao and Yang, Jingkang and Zhang, Yuanhan and Wang, Ziyue and Tan, Haoran and Li, Chunyuan and Liu, Ziwei},
  journal={arXiv preprint arXiv:2406.16852},
  year={2024}
}

@misc{liu2024llavanext,
  title={Llavanext: Improved reasoning, ocr, and world knowledge},
  author={Liu, Haotian and Li, Chunyuan and Li, Yuheng and Li, Bo and Zhang, Yuanhan and Shen, Sheng and Lee, Yong Jae},
  year={2024}
}

@article{zhang2026hermes,
  title={HERMES: KV Cache as Hierarchical Memory for Efficient Streaming Video Understanding},
  author={Zhang, Haowei and Yang, Shudong and Fu, Jinlan and Ng, See-Kiong and Qiu, Xipeng},
  journal={arXiv preprint arXiv:2601.14724},
  year={2026}
}

@article{zhao2025cogstream,
  title={CogStream: Context-guided Streaming Video Question Answering},
  author={Zhao, Zicheng and Wang, Kangyu and Li, Shijie and Qian, Rui and Lin, Weiyao and Liu, Huabin},
  journal={arXiv preprint arXiv:2506.10516},
  year={2025}
}

@inproceedings{li2025lion,
  title={Lion-fs: Fast \& slow video-language thinker as online video assistant},
  author={Li, Wei and Hu, Bing and Shao, Rui and Shen, Leyang and Nie, Liqiang},
  booktitle={Proceedings of the IEEE/CVF Conference on Computer Vision and Pattern Recognition},
  pages={3240--3251},
  year={2025}
}

@article{qian2024streaming,
  title={Streaming long video understanding with large language models},
  author={Qian, Rui and Dong, Xiaoyi and Zhang, Pan and Zang, Yuhang and Ding, Shuangrui and Lin, Dahua and Wang, Jiaqi},
  journal={Advances in Neural Information Processing Systems},
  volume={37},
  pages={119336--119360},
  year={2024}
}

@article{chen2025streamkv,
  title={Streamkv: Streaming video question-answering with segment-based kv cache retrieval and compression},
  author={Chen, Yilong and Bai, Xiang and Wang, Zhibin and Bai, Chengyu and Dai, Yuhan and Lu, Ming and Zhang, Shanghang},
  journal={arXiv preprint arXiv:2511.07278},
  year={2025}
}

@article{tian2025ego,
  title={Ego-r1: Chain-of-tool-thought for ultra-long egocentric video reasoning},
  author={Tian, Shulin and Wang, Ruiqi and Guo, Hongming and Wu, Penghao and Dong, Yuhao and Wang, Xiuying and Yang, Jingkang and Zhang, Hao and Zhu, Hongyuan and Liu, Ziwei},
  journal={arXiv preprint arXiv:2506.13654},
  year={2025}
}

@inproceedings{chenscaling,
  title={Scaling RL to Long Videos},
  author={Chen, Yukang and Huang, Wei and Shi, Baifeng and Hu, Qinghao and Ye, Hanrong and Zhu, Ligeng and Liu, Zhijian and Molchanov, Pavlo and Kautz, Jan and QI, XIAOJUAN and others},
  booktitle={The Thirty-ninth Annual Conference on Neural Information Processing Systems}
}

@article{wang2025videothinker,
  title={Video-Thinker: Sparking" Thinking with Videos" via Reinforcement Learning},
  author={Wang, Shijian and Jin, Jiarui and Wang, Xingjian and Song, Linxin and Fu, Runhao and Wang, Hecheng and Ge, Zongyuan and Lu, Yuan and Cheng, Xuelian},
  journal={arXiv preprint arXiv:2510.23473},
  year={2025}
}

@article{dao2023flashattention,
  title={Flashattention-2: Faster attention with better parallelism and work partitioning},
  author={Dao, Tri},
  journal={arXiv preprint arXiv:2307.08691},
  year={2023}
}

@article{ye2025flashinfer,
  title={Flashinfer: Efficient and customizable attention engine for llm inference serving},
  author={Ye, Zihao and Chen, Lequn and Lai, Ruihang and Lin, Wuwei and Zhang, Yineng and Wang, Stephanie and Chen, Tianqi and Kasikci, Baris and Grover, Vinod and Krishnamurthy, Arvind and others},
  journal={Proceedings of Machine Learning and Systems},
  volume={7},
  year={2025}
}

@article{dong2024flex,
  title={Flex attention: A programming model for generating optimized attention kernels},
  author={Dong, Juechu and Feng, Boyuan and Guessous, Driss and Liang, Yanbo and He, Horace},
  journal={arXiv preprint arXiv:2412.05496},
  volume={2},
  number={3},
  pages={4},
  year={2024}
}

@article{hsu2024liger,
  title={Liger kernel: Efficient triton kernels for llm training},
  author={Hsu, Pin-Lun and Dai, Yun and Kothapalli, Vignesh and Song, Qingquan and Tang, Shao and Zhu, Siyu and Shimizu, Steven and Sahni, Shivam and Ning, Haowen and Chen, Yanning},
  journal={arXiv preprint arXiv:2410.10989},
  year={2024}
}

@article{yuan2025tarsier2,
  title={Tarsier2: Advancing large vision-language models from detailed video description to comprehensive video understanding},
  author={Yuan, Liping and Wang, Jiawei and Sun, Haomiao and Zhang, Yuchen and Lin, Yuan},
  journal={arXiv preprint arXiv:2501.07888},
  year={2025}
}

\newpage
\appendix
\section{More Implementation Details}

\subsection{Streaming Inference}
As illustrated in \cref{alg:inference}, recording both decoding
and pruning operations into static CUDA graphs eliminates per-step
kernel launch overhead while preserving explicit control over KV cache
manipulation. This design enables flexible KV cache updates required by
RCSM while achieving decoding throughput comparable to optimized
inference engines, making it suitable for high-throughput RL rollouts
and real-time streaming deployment. 

\begin{algorithm}[htbp]
\caption{Streaming Inference with CUDA Graph KV Cache Eviction}
\label{alg:inference}
\begin{algorithmic}[1]
\REQUIRE Video stream $\mathcal{V}$, Instruction $\mathcal{I}$, Window size $W$, Max new tokens $N$
\STATE Initialize $KV\_Cache \gets \emptyset$, $\text{Active\_Chunks} \gets 0$
\STATE Graph capture \textsc{DecodeKernel} and \textsc{EvictKernel}
\FOR{each video chunk $v_t \in \mathcal{V}$}
    \IF{$\text{Active\_Chunks} \ge W$}
        \STATE \COMMENT{In-place memory shift via captured CUDA Graph}
        \STATE \textsc{ReplayGraph}(\textsc{EvictKernel}, \text{target\_starts}, \text{source\_ends})
        \STATE $\text{Active\_Chunks} \gets \text{Active\_Chunks} - 1$
    \ENDIF
    \STATE \COMMENT{Eager execution for variable-length visual tokens}
    \STATE $logits, KV\_Cache \gets \textsc{Prefill}(KV\_Cache, v_t, \mathcal{I})$
    \FOR{$step = 1$ \TO $N$}
        \STATE \COMMENT{Zero-overhead decode via captured CUDA Graph}
        \STATE $logits, KV\_Cache \gets \textsc{ReplayGraph}(\textsc{DecodeKernel}, token_{step-1})$
        \STATE \COMMENT{Accelerated sampling via FlashInfer}
        \STATE $token_{step} \gets \textsc{Sample}(logits)$ 
        \IF{$token_{step} \in \{\langle\text{|im\_end|}\rangle\}$}
            \STATE \textbf{break}
        \ENDIF
    \ENDFOR
    \STATE $\text{Active\_Chunks} \gets \text{Active\_Chunks} + 1$
\ENDFOR
\end{algorithmic}
\end{algorithm}

To further optimize memory access and compute efficiency, our attention mechanism employs a static KV cache and integrates the \texttt{flash attn with kvcache} interface~\cite{dao2023flashattention}. Additionally, during the autoregressive generation phase, we leverage \texttt{top\_k\_top\_p\_sampling\_from\_logits} operator 
in FlashInfer~\cite{ye2025flashinfer} to accelerate the sampling process, thereby significantly minimizing the latency of token selection.

\subsection{Training}
we utilize FlexAttention~\cite{dong2024flex} to implement RCSM during the training phase. This allows us to flexibly define and customize the attention masks without incurring prohibitive computational overhead. Furthermore, we incorporate the Liger Kernel~\cite{hsu2024liger} to accelerate the training process and reduce peak GPU memory consumption, enabling more efficient training over extended video sequences.

\section{Dataset Information}
\label{sec:dataset_appendix}

\paragraph{\textbf{Video Sources.}}
To construct our highly diverse dataset, we sampled high-quality video subsets from the widely used LLaVA-Video-178K~\cite{zhang2024llava} and Tarsier2-Recap-585K~\cite{yuan2025tarsier2} datasets. These subsets provide a rich foundation of dynamic visual content, ensuring robust training for streaming video understanding.

\paragraph{\textbf{Diverse Instruction Synthesis Details.}}
As introduced in the main text, we synthesize instruction data by computing the Cartesian product of three key scenario dimensions to derive 39 meaningful combinations. The specific definitions and scope of these dimensions are detailed below:

\begin{itemize}
    \item \textbf{Interaction Modes:} This dimension defines the fundamental mechanism of how the user queries the streaming assistant.
    \begin{itemize}
        \item Real-time Dialogue: The user asks real-time or retrospective questions where the question and answer share the exact same timestamp. Formats include open-ended, multiple choice, binary, and counting questions.
        \item Event Trigger: The user sets a monitoring or predictive rule significantly in advance. The assistant must remain silent for a long duration until a specific event occurs or a prediction condition is met, at which point it triggers an alert.
        \item Continuous Output: The user requests a continuous stream of updates. The assistant generates multiple consecutive messages to dynamically describe ongoing changes, actions, or events as they unfold over time.
    \end{itemize}

    \item \textbf{Temporal Scopes:} This dimension specifies the time frame the instruction targets relative to the current timestamp.
    \begin{itemize}
        \item Past: Retrospective queries requiring the assistant to recall details about events that occurred in the distant past, relying heavily on long-term memory retrieval.
        \item Current: Real-time queries focusing on the immediate present context regarding elements currently visible in the video stream.
        \item Future: Predictive queries asking for future forecasts or guidance on next steps concerning what is likely to happen based on current visual evidence.
    \end{itemize}

    \item \textbf{Content Semantics:} We comprehensively cover seven fine-grained visual and logical dimensions to ensure thorough video understanding:
    \begin{itemize}
        \item Entity \& Attribute Perception: Recognizing objects and their detailed visual attributes (e.g., color, shape, type).
        \item Action \& Activity Semantics: Identifying specific actions, behaviors, or ongoing activities.
        \item Spatial \& Geometric Relationships: Understanding spatial positions, distances, or geometric arrangements of objects.
        \item Causal \& Logical Reasoning: Deducing the underlying causes, reasons, or logical implications of observed events.
        \item Procedural State \& Evolution: Tracking the progress, steps, or current state of a structured procedure.
        \item Global Scene \& Context: Recognizing the overall environment, location type, weather, or scene atmosphere.
        \item Optical Character Recognition (OCR): Extracting text content, signage, or characters visible on the screen.
    \end{itemize}
\end{itemize}

\begin{figure*}[htbp]
    \centering
    \begin{minipage}[b]{0.43\textwidth}
        \centering
        \includegraphics[width=\textwidth]{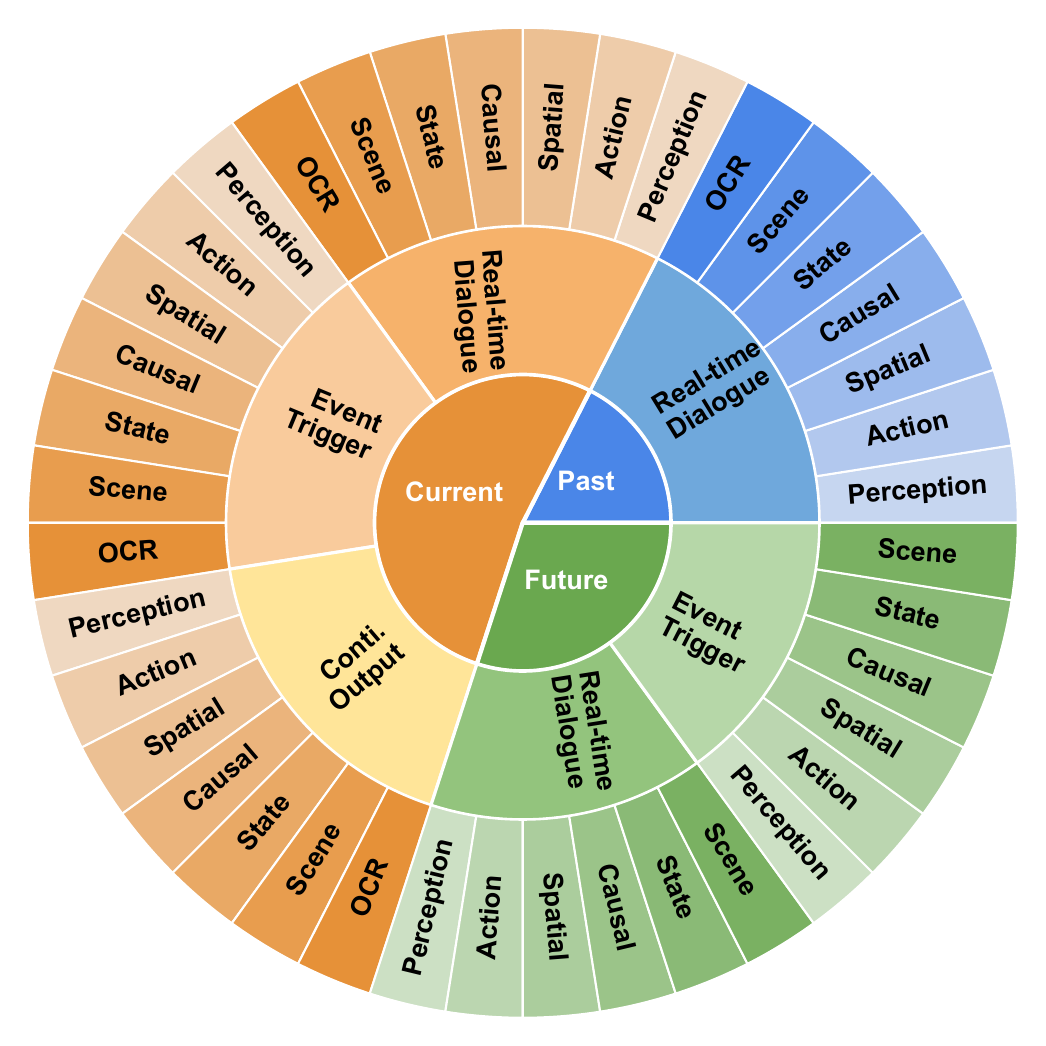}
        \subcaption{Data Distribution}
        \label{fig:data_dist}
    \end{minipage}
    \hfill
    \begin{minipage}[b]{0.55\textwidth}
        \centering
        \includegraphics[width=\textwidth]{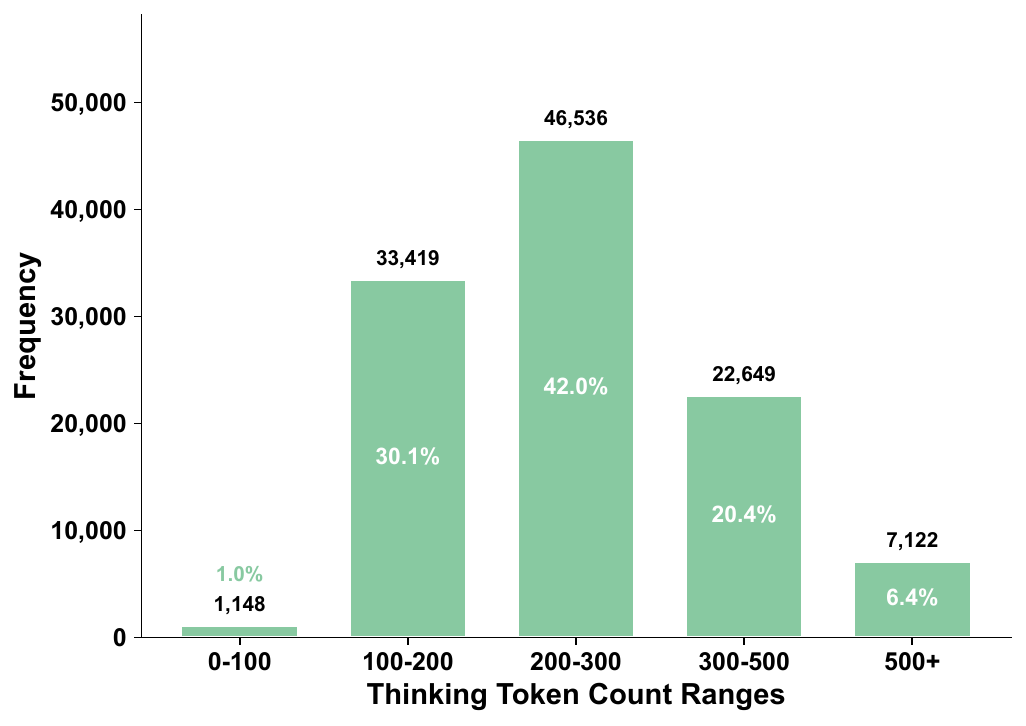}
        \subcaption{Thinking Token Length}
        \label{fig:token_len}
    \end{minipage}
    \caption{\textbf{Statistics of our ThinkStream Dataset.} We show (a) the hierarchical composition of data and (b) the distribution of thinking process lengths.}
    \label{fig:overall_stats}
\end{figure*}

\paragraph{\textbf{Dataset Statistics and Analysis.}}
\cref{fig:overall_stats} illustrates the statistical distribution of our constructed ThinkStream dataset. As shown in the figure, our dataset exhibits a rich diversity across hierarchical compositions—encompassing various interaction modes, temporal scopes, and fine-grained content semantics—as well as a robust distribution of thinking token lengths. This comprehensive variety ensures that the model is exposed to a wide range of streaming video understanding scenarios during training.

\section{More Results}

To further demonstrate the scalability and effectiveness of our proposed approach, we trained our model based on the Qwen2.5-VL-7B~\cite{Qwen2.5-VL} architecture, denoted as ThinkStream-7B. As shown in \cref{tab:ovo_more}, ThinkStream-7B achieves significant performance improvements over the base Qwen2.5-VL-7B model across multiple dimensions on the OVO-Bench streaming video benchmark. 

Specifically, in the OCR evaluation, ThinkStream-7B attains an impressive score of 87.92, outperforming the base model's 67.79 by over 20 points. In terms of Real-Time Visual Perception, our model improves the average score from 59.90 to 69.12. Furthermore, ThinkStream-7B demonstrates enhanced capability in long-term memory retrieval, boosting the average score of Backward Tracing from 52.28 to 60.68. These substantial gains indicate that integrating our streaming framework into stronger base models yields consistent and robust enhancements for streaming video understanding.

\begin{table*}[htbp]
    \centering
    \caption{Performance comparison of ThinkStream and existing open-source offline and online models on the OVO-Bench streaming video benchmark.}
    \label{tab:ovo_more}
    \resizebox{\textwidth}{!}{
    \begin{tabular}{l | cccccc | c | ccc | c | c}
        \toprule
        \multirow{2}{*}{\textbf{Model}} & \multicolumn{7}{c|}{\textbf{Real-Time Visual Perception}} & \multicolumn{4}{c|}{\textbf{Backward Tracing}} & \multirow{2}{*}{\textbf{Avg.}} \\
        \cline{2-12}
        & OCR & ACR & ATR & STU & FPD & OJR & Avg. & EPM & ASI & HLD & Avg. & \\
        \midrule
        \multicolumn{13}{c}{\textbf{Open-source Offline Models}} \\
        \midrule
        LLaVA-Video-7B~\cite{zhang2024llava} & 69.13 & 58.72 & 68.83 & 49.44 & 74.26 & 59.78 & 63.52 & 56.23 & 57.43 & 7.53 & 40.4 & 51.96 \\
        Qwen2-VL-7B~\cite{wang2024qwen2} & 60.4 & 50.46 & 56.03 & 47.19 & 66.34 & 55.43 & 55.98 & 47.81 & 35.48 & 56.08 & 46.46 & 51.22 \\
        LongVU-7B~\cite{shen2024longvu} & 53.69 & 53.21 & 62.93 & 47.75 & 68.32 & 59.78 & 57.61 & 40.74 & 59.46 & 4.84 & 35.01 & 46.31 \\
        Qwen2.5-VL-3B~\cite{Qwen2.5-VL} & 76.51 & 44.03 & 67.24 & 42.13 & 68.31 & 61.96 & 60.03 & 50.50 & 53.38 & 22.04 & 41.98 & 51.00 \\
        Qwen2.5-VL-7B~\cite{Qwen2.5-VL} & 67.79 & 55.05 & 67.24 & 42.13 & 66.34 & 60.87 & 59.90 & 51.52 & 58.78 & 23.66 & 44.65 & 52.28 \\
        Qwen2.5-VL-32B~\cite{Qwen2.5-VL} & 77.18 & 58.72 & 68.10 & 50.56 & 74.26 & 57.61 & 64.40 & 58.59 & 62.84 & 29.57 & 50.33 & 57.37 \\
        \midrule
        \multicolumn{13}{c}{\textbf{Open-source Online Models}} \\
        \midrule
        Flash-VStream-7B~\cite{zhang2025flash} & 24.16 & 29.36 & 28.45 & 33.71 & 25.74 & 28.8 & 28.37 & 39.06 & 37.16 & 5.91 & 27.38 & 27.88 \\
        VideoLLM-online-8B~\cite{chen2024videollm} & 8.05 & 23.85 & 12.07 & 14.04 & 45.54 & 21.2 & 20.79 & 22.22 & 18.8 & 12.18 & 17.73 & 19.26 \\
        Dispider-7B~\cite{qian2025dispider} & 57.72 & 49.54 & 62.07 & 44.94 & 61.39 & 51.63 & 54.55 & 48.48 & 55.41 & 4.3 & 36.06 & 45.31 \\
        Streamo-3B~\cite{xia2025streaming} & 78.52 & 52.29 & 67.24 & 44.38 & 55.45 & \textbf{71.20} & 61.51 & 51.18 & 57.43 & 16.67 & 41.76 & 51.64 \\
        StreamForest-7B~\cite{zeng2025streamforest} & 68.46 & 53.21 & 71.55 & 47.75 & 65.35 & 60.87 & 61.20 & \textbf{58.92} & \textbf{64.86} & 32.26 & 52.02 & 56.61 \\
        \tableLineColor \textbf{ThinkStream-3B (Ours)} & \underline{85.23} & \textbf{64.22} & 69.82 & \underline{49.43} & \underline{69.31} & 64.13 & \underline{67.03} & \underline{53.87} & 59.46 & \underline{43.55} & \underline{52.30} & \underline{59.66} \\
        \tableLineColor \textbf{ThinkStream-7B (Ours)} & \textbf{87.92} & \underline{59.63} & \textbf{75.00} & \textbf{53.93} & \textbf{70.30} & \underline{67.93} & \textbf{69.12} & 51.17 & \underline{64.19} & \textbf{66.67} & \textbf{60.68} & \textbf{64.90} \\
        \bottomrule
    \end{tabular}
    }
\end{table*}

\section{Qualitative Analysis}
\begin{figure*}[htbp]
    \includegraphics[width=\textwidth]{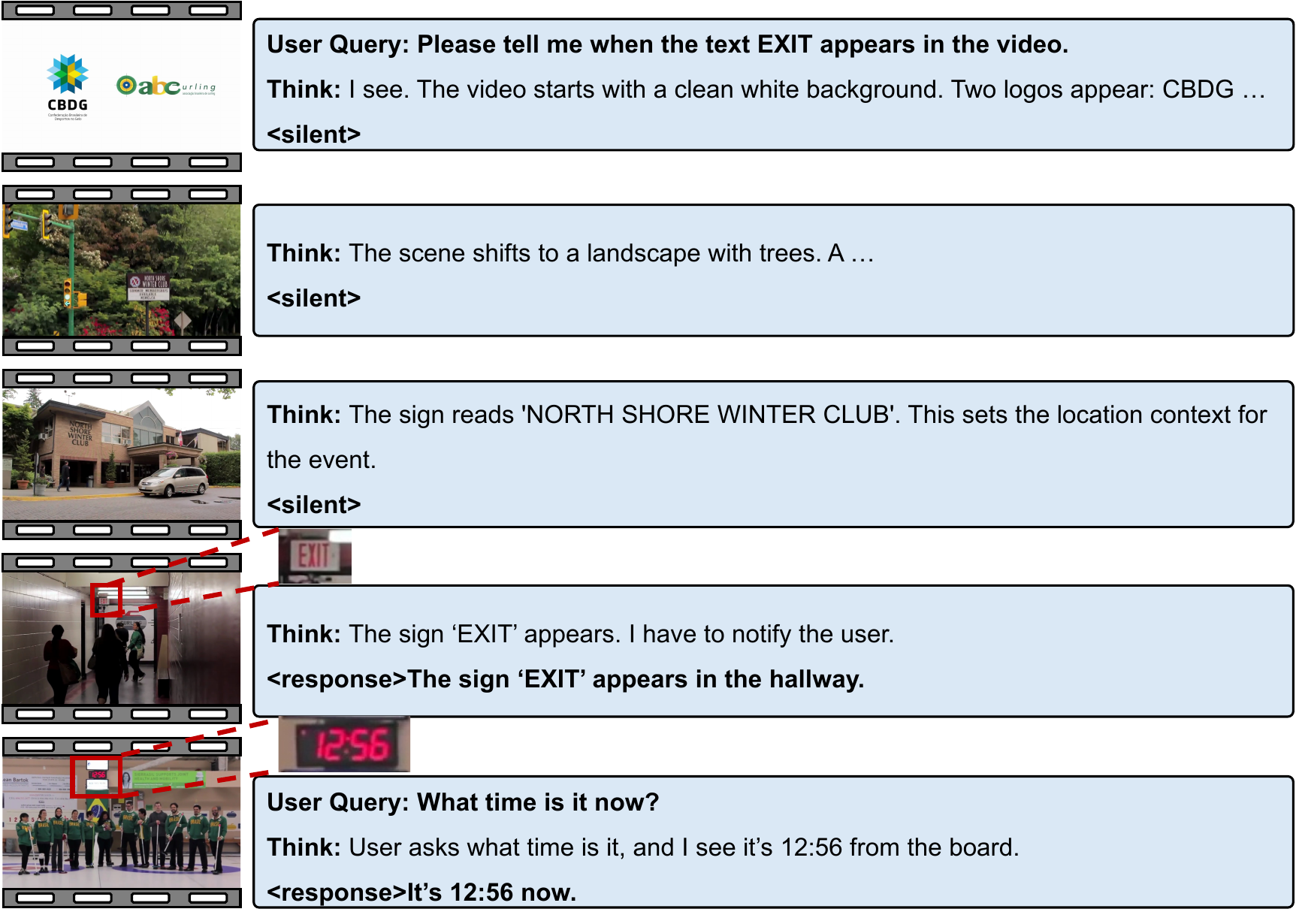}
    \caption{Qualitative Example}
    \label{fig:example}
\end{figure*}

\cref{fig:example} illustrates the Watch-Think-Speak paradigm across two scenarios. First, a user sets an event trigger for the text ``EXIT.'' The model leverages an incremental reasoning trace (\texttt{<think>}), correctly outputting \texttt{<silent>} during irrelevant scenes until the target appears, prompting a \texttt{<response>}. Second, a real-time query asks for the current time, which the model instantly identifies from visual evidence to deliver an accurate answer. These examples demonstrate the framework's ability to seamlessly unify perception and reasoning, autonomously deciding when to observe silently and when to interact without relying on external classifiers.

\section{Prompt Used}

For the synthesis of ThinkStream instruction data, we design a specialized prompt that strictly relies on provided video segment descriptions. The complete prompt for the instruction data generation is presented as follows:

\begin{center}
\begin{tcolorbox}[
    breakable, 
    colback=white,        
    colframe=black,       
    arc=0pt, outer arc=0pt, 
    boxrule=0.4pt,        
    width=0.95\linewidth  
]
\small
\textbf{System:} You are an expert in creating training data for Video-LLMs, specifically focusing on Streaming Video Q\&A. Your task is to generate a JSON dialogue history strictly based on the provided descriptions and configurations. Follow these strict requirements: \\[0.5em]
1. Content Constraints: Base all dialogue exclusively on the provided video descriptions. Do not hallucinate details (actions, objects, text) not explicitly present. \\
2. Message Quantity: The generated dialogue history must contain exactly ONE User message. \\
3. Timing \& Length Constraints: Every single Assistant message should be under 20 words. If multiple Assistant messages are generated, the time gap between consecutive Assistant messages should be at least 3.0 seconds (does NOT apply between a User and an Assistant message). The Assistant's first response must occur after the User's trigger. \\
4. Chronology: Timestamps must be floats (in seconds) strictly within segment boundaries and in chronological order. \\
5. Relative Temporal References: Strictly avoid using absolute timestamps (e.g., 'at 3.5 seconds') in the dialogue content. Use relative temporal markers based on events or actions (e.g., 'When the car turns left...'). \\
6. Strict Compliance: You must strictly follow all requirements defined in the Scenario Configuration and Response Format Configuration. \\[0.5em]
\textbf{User:} \\
INPUT DATA: \\
Video Segment Descriptions: \{input\_data\} \\
Scenario Configuration: \{scenario\_description\} \\
Response Format Configuration: \{format\_instruction\} \\[0.5em]
STRICT OUTPUT FORMAT: \\
Return a single JSON list containing the dialogue history (return \texttt{[]} if no valid dialogue can be generated): \\
\texttt{[ \{{"role": "user", "content": "<generated\_query>", "timestamp": <float\_seconds>}\} ]}
\end{tcolorbox}
\end{center}

To generate the internal reasoning trace of the model during real-time processing, we employ a Chain-of-Thought (CoT) data synthesis prompt. The detailed prompt for the CoT annotation task is shown below:

\begin{center}
\begin{tcolorbox}[
    breakable, 
    colback=white,        
    colframe=black,       
    arc=0pt, outer arc=0pt, 
    boxrule=0.4pt,        
    width=0.95\linewidth  
]
\small
\textbf{System:} You are a data synthesis engine for a Real-Time Streaming Multimodal AI. Your task is to generate an "Internal Stream of Consciousness" (Chain-of-Thought) based on video captions and user-assistant dialogue. Follow these strict requirements: \\[0.5em]
1. Temporal Causality (No Future Peeking): A thought at timestamp \texttt{T\_think} can ONLY reference captions or user messages where \texttt{timestamp $\le$ T\_think}. You strictly cannot mention events or user questions that happen after the thought's timestamp. \\
2. Timing \& Density: The minimum gap between consecutive thoughts should be at least 2.0 seconds. Even when the user hasn't asked a question, continuously output "think" content. The generated CoT should extend to \texttt{min(last\_conversation\_timestamp + 10s, last\_caption\_end\_time)}. \\
3. User Query Response: When a user asks a question at timestamp \texttt{T\_user}, you MUST generate a thought at exactly the same timestamp \texttt{T\_user} that directly addresses the user's question, then resume continuous thinking. \\
4. Length: Every single "think" content should be no more than 50 words. \\[0.5em]
\textbf{User:} \\
INPUT CONTEXT: \\
Time-Grounded Captions: \\
\{captions\} \\[0.5em]
User-Assistant Conversation: \\
\{conversation\} \\[0.5em]
STRICT OUTPUT FORMAT: \\
Return ONLY a valid JSON list of objects in ascending order of timestamp: \\
\texttt{[ \{{"think": "<think\_content>", "timestamp": <float\_seconds>}\} ]}
\end{tcolorbox}
\end{center}

\end{document}